\documentclass[runningheads]{llncs}

\usepackage[ruled,linesnumbered,algo2e,vlined]{algorithm2e}
\usepackage{amsfonts,amssymb}
\usepackage{booktabs}
\usepackage{caption}
\usepackage{comment}
\usepackage{graphicx}
\usepackage{hyperref}
\usepackage{paralist}
\usepackage{soul}
\usepackage{subcaption}
\usepackage{tikz}
\usepackage{url}
\usepackage[colorinlistoftodos,textsize=footnotesize]{todonotes}
\usepackage{wrapfig}

\usepackage{dcr}

\SetCommentSty{mycommfont}
\newlength\mylen

\SetKwComment{Comment}{$\triangleright$\ }{}

\begin{document}
    \title{
    A Monitoring and Discovery Approach for Declarative Processes Based on Streams}
    \titlerunning{A Discovery Approach for Declarative Processes Based on Streams}
    \author{
        Andrea Burattin \inst{1} \and
        Hugo A. L\'opez \inst{2} \and
        Lasse Starklit \inst{3}\thanks{Alphabetical order, equal authors contribution}}
    \authorrunning{Burattin et al.}

    \institute{Technical University of Denmark, Kgs. Lyngby, Denmark \and University of Copenhagen, Copenhagen, Denmark \and Netcompany A/S, Copenhagen, Denmark }
    \maketitle

\begin{abstract}
    Process discovery is a family of techniques that helps to comprehend processes from their data footprints. Yet, as processes change over time so should their corresponding models, and failure to do so will lead to models that under- or over-approximate behavior. We present a discovery algorithm that extracts declarative processes as Dynamic Condition Response (DCR) graphs from event streams. Streams are monitored to generate temporal representations of the process, later processed to generate declarative models. We validated the technique via quantitative and qualitative evaluations. For the quantitative evaluation, we adopted an extended Jaccard similarity measure to account for process change in a declarative setting. For the qualitative evaluation, we showcase how changes identified by the technique correspond to real changes in an existing process. The technique and the data used for testing are available online.

\end{abstract}

\section{Introduction}

The only constant aspect of processes is their change. Either because of internal organization restructuring or because of variables external to the organization, processes are required to adapt quickly to achieve required outcomes. The COVID-19 pandemic showed us how organizations needed to move from physical work to hybrid or remote production facilities, forcing them to abandon optimized routes toward new flows. In administrative processes, each regulation change will require municipal governments to adapt their processes to preserve compliance. In Denmark, the laws determining the guidelines for case management in the social sector had 4,686 changes between 2009 and 2020~\cite{justitiaReport}.

Process mining approaches promise that given enough data, a control-flow discovery technique will generate a model that is as close to reality as possible. This evidence-based approach has a caveat: one needs to assume that the observations that were used as inputs belong to the same process. Not taking into consideration change might end in under- or over-constrained processes that do not represent the reality of the process. The second assumption is that it is possible to identify complete traces from the event log. This requirement indeed presents considerable obstacles in the organizations in which processes are constantly happening and evolving, either because the starting events are located in legacy systems no longer in use, or because current traces have not finished yet. 

Accounting for change is particularly important in declarative processes. Based on a ``outside-in'' approach, declarative processes describe the minimal set of rules that generate accepting traces. To achieve this goal, declarative models place constraints between activities such that they restrict or enforce only compliant behavior. For process mining, the simplicity of declarative processes has been demonstrated to fit well with real process executions, and declarative miners are at the moment the most precise miners in use\footnote{See \url{https://icpmconference.org/2021/process-discovery-contest/}.}. However, little research exists regarding how declarative miners are sensitive to process change.

The objective of this paper is to study how declarative miners can give accurate and timely views of incomplete traces (so-called \emph{event streams}). 
We integrate techniques of streaming  process mining to declarative modeling notations, in particular, DCR graphs~\cite{hildebrandt_declarative_2010}. While previous techniques of streaming conformance checking have been applied to other declarative languages (e.g.: Declare~\cite{pesic2006declarative}), these languages are fundamentally different. Declare provides a predefined set of 18 constraint templates taking inspiration from~\cite{dwyer1999patterns} with an underlying semantics based on LTL formulae on finite traces~\cite{de2014reasoning}. Instead, DCR is based on a minimal set of 5 constraints, being able to capture regular and omega-regular languages~\cite{DBLP:journals/acta/DeboisHS18}. The choice of DCR is not fortuitous: DCR is an industrial process language integrated into KMD Workzone, a case management solution used by 70\% of central government institutions in Denmark~\cite{DBLP:conf/bpm/NorgaardAMDLJ17}. For the metrics, we present a fast \& syntax-driven model-to-model metric and show its suitability in quantifying model changes.

Event streams present challenges for process discovery. Streams are potentially infinite, making memory and time computation complexities major issues. Our technique optimizes these aspects by relying on intermediate representations that will be updated at runtime. Another aspect considered is \emph{extensibility}: our techniques not only rely on the minimal set of DCR constraints but it can be extended to more complex workflow patterns that combine atomic constraints. 

\begin{figure}
    \centering
    \includegraphics[width=\textwidth]{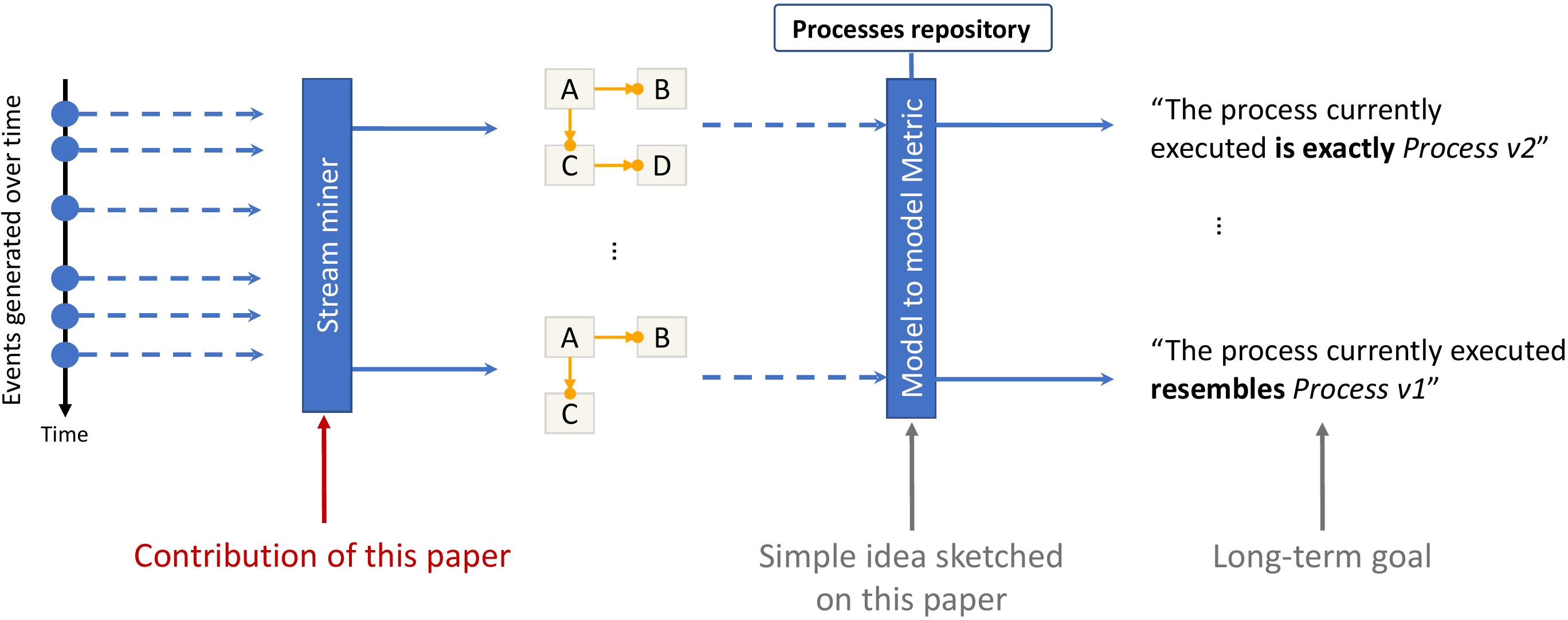}
    \caption{Contribution of the paper}
    \label{fig:contribution}
    \vspace{-1em}
\end{figure}
Fig.~\ref{fig:contribution} shows the paper's contribution: a streaming mining component, capable of continuously generating DCR graphs from an event stream (here we use the plural \textit{graphs} to indicate that the DCR model could evolve over time, to accommodate drifts in the model that might occur). Towards the long-term goal of a system capable of spotting changes in a detailed fashion, we will also sketch a simple model-to-model metric for DCR, which can be used to compare the results of the stream mining with a catalog or repository of processes. An implementation of our techniques is available in Java and it can be downloaded, together with all the tests and datasets\footnote{See \url{https://github.com/beamline/discovery-dcr}.}.

The rest of the paper is structured as follows:
related works are presented in Section~\ref{sec:relwork};
background on streaming process discovery and DCR graphs is in Section~\ref{sec:background}.
The streaming discovery is presented in Section~\ref{sec:onlinediscovery} 
and the approach is validated in Section~\ref{sec:validity}.
Section~\ref{sec:conclusions} concludes the paper.

\section{Related Work} \label{sec:relwork}

This paper is the first work aiming at the discovery of DCR graphs from an event stream. In the literature, it is possible to find work referring to either offline discovery for DCR graphs or online discovery for Declare models. In the rest of the section, we will also discuss streaming process mining in general terms.

\textit{Offline process discovery techniques.}
The state of the art in the offline discovery of DCR is represented by the DisCoveR algorithm~\cite{Back2021}. In their paper, the authors claim an accuracy of 96,2\% with linear time complexity. 
The algorithm is an extension of the ParNek algorithm~\cite{nekrasaite2019discovering} but it uses a highly efficient implementation of DCR, mapping to bit vectors, where each activity corresponds to a particular index of the vector. 
 A more recent approach described in~\cite{DBLP:conf/bpm/SlaatsDB21}, presents the Rejection miner which exploits the idea of having both positive and negative examples to produce a better process model.

Related to what we present in this paper are conformance checking~\cite{Carmona2018} and process repair techniques~\cite{VanderAalst2016}. Both these fields aim at understanding whether executions can be replayed on top of an existing processes model or not. However, in our case, we wanted to separate the identification of the processes (i.e., control-flow discovery) from the calculation of their similarity (i.e., the model-to-model metric) so that these two contributions can be used independently from each other. Conformance checking and process repair, on the other hand, embed the evaluation and the improvement into one ``activity''.

\textit{Online Discovery for Declarative Models.} 
In~\cite{burattin2015online} a framework for the discovery of Declare models from streaming event data has been proposed. This framework can be used to process events online, as they occur, as a way to deal with large and complex collections of datasets that are impossible to store and process altogether. In~\cite{9207500} the work has been generalized to handle also the mining of data constraints, leveraging the MP-Declare notation~\cite{BurattinMS16}.

\textit{Streaming Process Mining in General.}
Another important research conducted within the area of online process mining is the work done by van Zelst, in his Ph.D. thesis~\cite{VanZelst2019}. Throughout the thesis, the author proposes process mining techniques applicable to process discovery, conformance checking, and process enhancement from event streams. An important conclusion from his research consists of the idea of building intermediate models that can capture the knowledge observed in the stream before creating the final process model.

In~\cite{DBLP:reference/bdt/Burattin19} the author presents a taxonomy for the classification of streaming process mining techniques. Our techniques constitute a hybrid approach in the categories in ~\cite{DBLP:reference/bdt/Burattin19},  mixing a smart window-based model which is used to construct and maintain an intermediate structure updated, and a problem reduction technique used to transform such structure into a DCR graph.

\section{Background} \label{sec:background}

In the following section, we recall basic notions of Directly Follows Graphs~\cite{vanderAalst2016new} and  the Dynamic Condition Response (DCR)
graphs~\cite{hildebrandt_declarative_2010}. While, in general, DCR  is expressive enough to capture multi-perspective constraints such as time~\cite{DBLP:journals/jlp/HildebrandtMSZ13}, data~\cite{DBLP:conf/bpm/StromstedLDM18}, sub-processes~\cite{FM15} and message-passing constraints~\cite{DCRChoreo19},  in this paper we use the classical, set-based formulation first presented in~\cite{hildebrandt_declarative_2010} that contains only four most basic behavioural relations: conditions, responses, inclusions and exclusions.
%

\begin{definition}[Sets, Events and Sequences]
    Let $\mathcal{C}$ denote the set of possible case identifiers and let $\mathcal{A}$ denote the set of possible activity names. 
    The event universe is the set of all possible events $\mathcal{E} = \mathcal{C} \times \mathcal{A}$ and an event is an element $e =(c,a) \in \mathcal{E}$.
    Given a set $\mathbb{N}_n^+ = 1, 2, \dots, n$ and a target set $A$, a sequence $\sigma : \mathbb{N}_n^+ \to A$ maps index values to elements in $A$. For simplicity we can consider sequences using a string interpretation: $\sigma = \langle a_1,\dots,a_n \rangle$ where $\sigma(i) = a_i \in A$.
\end{definition}
With this definition, we can now formally characterize an event stream:
\begin{definition}[Event stream]
    An event stream is an unbounded sequence mapping indexes to events: $\mathcal{S}: \mathbb{N}^+ \to \mathcal{E}$.
\end{definition}
Our approach for extracting DCR graphs leverages the notion of Extended Directly Follows Grap, which is an extension of DFG:
\begin{definition}[Directly Follows Graph (DFG)] \label{def:dfg}
    A directly follows graph is a graph $G = (V,R)$ where nodes represent activities (i.e., $V \subseteq \mathcal{A}$), and edges indicate directly follows relations from source to  target activities (i.e., $(a_s,a_t) \in R$ with $a_s, a_t \in V$, so $R \subseteq V \times V$).
\end{definition}

\begin{definition}[Extended DFG] \label{def:extended-dfg}
    An extended DFG is a graph $G_x = (V, R, X)$ where $(V,R)$ is a DFG and $X$ contains additional numerical attributes referring to the nodes: $X: V \times \textit{Attrs} \to \mathbb{R}$, where $\textit{Attrs}$ is the set of all attribute names. To access attribute $a_1$ for node $v$ we use the notation $X(v,a_1)$.
\end{definition}
In the rest of the paper we will consider the following attributes: 
    \textsf{avgFO}: average first occurrence of the activity among the traces seen so far;
    \textsf{noTraceApp}: number of traces that the activity has appeared in so far;
    \textsf{avgIdx}: average occurrence index of the activity; and
    \textsf{noOccur}: number of occurrences of the activity.
A DCR graph consists of a multi-directed graph and a marking. 
\begin{definition}[DCR Graph] \label{def:dcr}
    A \emph{DCR graph} is a tuple $\langle \mathcal{A},M,
    \conditionrel,\responserel,\includerel,\excluderel \rangle$, where
     $\mathcal{A}$ is a set of activities,  
    $M \in {\cal{P}}(\mathcal{A}) \times {\cal{P}}(\mathcal{A}) \times {\cal{P}}(\mathcal{A})$
    is a \emph{marking}, 
    and $\phi\subseteq \mathcal{A}\times \mathcal{A}$ for $\phi\in \reltypesWithoutMilestone$ are \emph{relations} between activities.
\end{definition}

A DCR graph defines processes whose executions are finite and infinite sequences of activities. An activity may be executed several times. The three sets of activities in the marking  $M=(\Ex,\Re,\In)$ define the state of a process, and they are referred to as the \emph{executed} activities ($\Ex$), the \emph{pending} response ($\Re$)\footnote{We might simply say pending when it is clear from the context.} and the \emph{included} activities ($\In$). DCR relations define what is the effect of executing one activity in the graph for its context. Briefly:
\begin{itemize}
    \item Condition relations $a\conditionrel a'$ say that the execution of $a$ is a prerequisite for $a'$, i.e. if $a$ is included, then $a$ must have been executed for $a'$ to be enabled for execution. 
    \item Response relations $a\responserel a'$ say that whenever $a$ is executed, $a'$ becomes pending. In a run, a pending event must eventually be executed 
    or be excluded. We refer to $a'$ as a response to $a$.
    \item An inclusion (respectively exclusion) relation $a\includerel a'$ (respectively $a\excluderel a'$) means that if $a$ is executed, then $a'$ is included (respectively excluded). 
\end{itemize}

For a DCR graph\footnote{We will use DCR graph and DCR model interchangeably in this paper} $P$ with activities $\mathcal{A}$ and marking $M=(\Ex,\Re,\In)$ we write $P_{\responserel}$ for the set of pairs $\{(x,y) ~|~ \{x,y\} \in \mathcal{A} ~\land~ (x,y) \in \responserel \}$ (similarly for any of the relations in $\phi$) and we write $P_\mathcal{A}$ for the set of activities. 
Notice that  Definition \ref{def:dcr} omits the existence of a set of labels and labelling function present in \cite{hildebrandt_declarative_2010}. This has a consequence in the set of observable traces:  Assume the DCR graph $G = \langle \{a,b\},\{a: (\bot, \bot,\top),b: (\bot, \bot,\top)\}, \emptyset, \emptyset, \emptyset,\emptyset \rangle$ as well as a set of labels $L=\{p\}$ and a labelling function $ l= \{(a,p),(b,p)\}$. A possible run of $G$ has the shape $\sigma = \langle p, p \rangle$, which can be generated from 1) two executions of $a$, 2) two executions of $b$ or 3) an interleaved execution of $a$ and $b$. By removing the labels from the events (or alternatively, assuming an injective surjective labelling function in \cite{hildebrandt_declarative_2010}), we assume that if two events share the same activity name, it is because there was a repetition of the event in the stream.

\section{Streaming DCR Miner} \label{sec:onlinediscovery}

\newcommand{\obs}{\textsf{obs}}
\newcommand{\deps}{\textsf{deps}}
    
\begin{figure}[t]
    \centering
    \includegraphics[width=\textwidth]{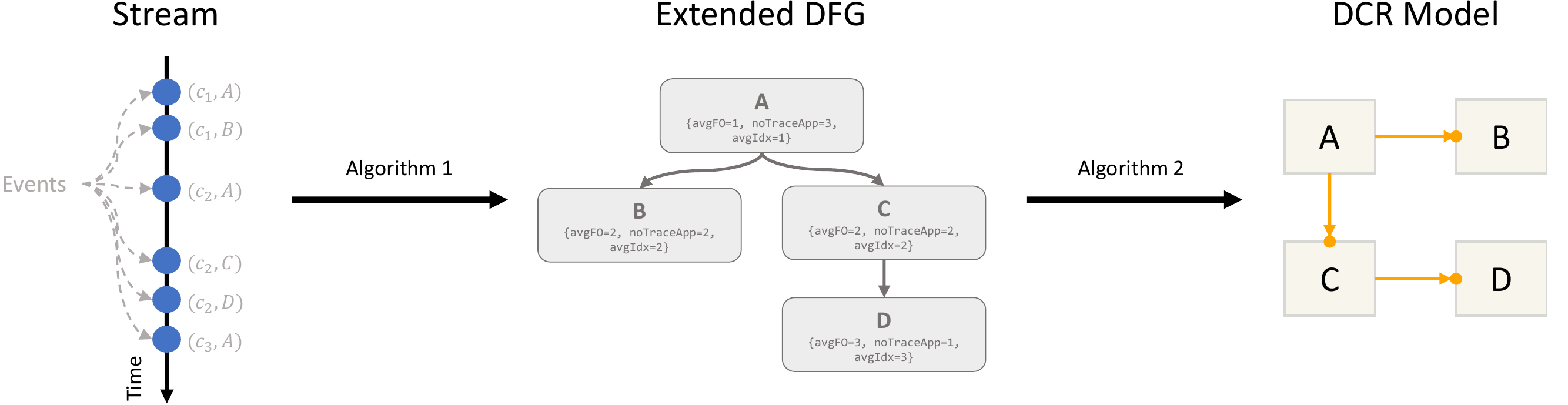}
    \caption{Conceptual representation of the discovery strategy in this paper.}
    \label{fig:idea}
    \vspace{-1em}
\end{figure}

This section presents the general structure of the stream mining algorithm for DCR graphs. The general idea of the approach presented in this paper is depicted in Fig.~\ref{fig:idea}: constructing and maintaining an extended DFG structure (cf. Def.~\ref{def:extended-dfg}) starting from the stream and then, periodically, a new DCR graph is extracted from the most recent version of the extended DFG available. The extraction of the different DCR rules starts from the same extended DFG instance.
For readability purposes, we split the approach into two phases. The former (Alg.~\ref{alg:meta}) is in charge of extracting the extended DFG, the latter (Algs.~\ref{alg:miner:pattern}, \ref{alg:miner:patternAtomic}, \ref{alg:miner:patternComposite}) focuses on the extraction of DCR rules from the extended DFG.

\begin{algorithm2e}[h!]
	\caption{General structure of Streaming DCR Miner \label{alg:meta}}
        \footnotesize
	\DontPrintSemicolon
	\KwIn{$S$: stream of events \newline
	    $m_t$: maximum number of traces to store \newline
	    $m_e$: maximum number of events per trace to store \newline
	    $\langle T, \leq \rangle$: Pattern poset
	    }
	\DontPrintSemicolon
	\BlankLine
	Initialize map \obs\ \Comment{Maps case ids to sequence of activities}
	Initialize map \deps\ \Comment{Maps case ids to one activity name}
	Initialize extended DFG $G_X = (V, R, X)$ \;
	\For{\hspace{-0.1cm}\emph{\textbf{ever}}} {
        \Comment{Step 0: Observe new activity $a$ for case $c$}
        $(c, a) \gets \mathit{observe}(S)$ \;
        \BlankLine
        \Comment{Step 1: Update of the extended DFG}
        \eIf{$c \in \obs$}{
            Refresh the update time of $c$ \;
            \If{$|\obs(c)| \geq m_e$} {
                Remove oldest event from list $obs(c)$ \;
                Update $V$ and $X$ of $G_X$ to be consistent with the event just removed \;
            }
        } {
            \If{$|\obs| \geq m_t$} {
                Remove the oldest trace from $\obs$ and all its events \;
                Update $V$ and $X$ of $G_X$ to be consistent with the events just removed \;
            }
            $\obs(c) \gets \langle \rangle$ \Comment{Create empty list for $\obs(c)$}
        }
        $\obs(c) \gets \obs(c) \cdot \langle a \rangle $ \Comment{Append $a$ to $\obs(c)$}
        $V\gets V \cup \{a\}$ \;
        Update frequency and avg appearance index in $X$ component of $G_X$ \Comment{The average appearance index is updated considering the new position given by $|obs(c)|$}
        \If {$c \in \deps$} {
            $R \gets R \cup \{ (\deps(c), a) \}$ \;
        }
        $\deps(c) \gets a$ \;

	    \BlankLine
	    \Comment{Step 2: Periodic update of the DCR model}
	    \If{trigger periodic update of the model} {
	        $M \gets \textsf{mine}(\langle T, \leq \rangle, G_X)$ \tcp{See Algorithm~\ref{alg:miner:pattern}}
	        Notify about new model $M$ \;
	    }
	}
\end{algorithm2e}

Algorithm~\ref{alg:meta} takes as input a stream of events $S$, two parameters referring to the maximum number of traces $m_t$ and events to store $m_e$ and a set of DCR patterns to mine (see Def.~\ref{def:patternPoset}).
The algorithm starts by initializing two supporting map data structures $\obs$ and $\deps$ as well as an empty extended DCR graph $G_X$ (lines 1-3). $\obs$ is a map associating case ids to sequences of partial traces; $\deps$ is a map associating case ids to activity names.
After initialization, the algorithm starts consuming the actual events in a never-ending loop (line 4). The initial step consists of receiving a new event (line 5).
Then, two major steps take place: the first step consists of updating the extended DFG graph; the second consists of transforming the extended DFG into a DCR model.
To update the extended DFG
the algorithm first updates the set of nodes and extra attributes. If the case id $c$ of the new event has been seen before (line 6), then the algorithm refreshes the update time of the case id (line 7, useful to keep track of which cases are the most recent ones) and checks whether the maximum length of the partial trace for that case id has been reached (line 8). If that is the case, then the oldest event is removed and the $G_X$ is updated to incorporate the removal of the event. If this is the first time this case id is seen (line 11), then it is first necessary to verify that the new case can be accommodated (line 12) and, if there is no room, then first some space needs to be created by removing oldest cases and propagating corresponding changes (lines 13-14) and then a new empty list can be created to host the partial trace (line 15). In either situation, the new event is added to the partial trace (line 16) and, if needed, a new node is added to the set of vertices $V$ (line 17). The $X$ data structure can be refreshed by leveraging the properties of the partial trace seen so far (line 18).
To update the relations in the extended DFG (i.e., the $R$ component of $G_X$), the algorithm checks whether an activity was seen previously for the given case id $c$ and, if that is the case, the relation from such activity (i.e., $\deps(c)$) to the new activity just seen (i.e., $a$) is added (lines 19-20). In any case, the activity just observed is now the latest activity for case id $c$ (line 21).
Finally, according to some periodicity (line 22), the algorithm refreshes the DCR model by calling the procedure that transforms (lines 23-24) the extended DFG into a DCR model (cf. Alg.~\ref{alg:miner:pattern}).

Algorithm \ref{alg:miner:pattern} generates a DCR graph from an extended DFG. We do so by (1) defining patterns that describe occurrences of atomic DCR constraints in the extended DFG, and (2) defining composite patterns from (1) that define the most common behavior. 

\begin{definition}[Pattern Poset]\label{def:patternPoset}
    Given a set of patterns $T$,  $\langle T, \leq \rangle$ is the pattern dependency poset where $\leq$ is a  (binary) relation over $T$ such that reflexivity $(x,x) \in \leq$, transitivity $(x,y) \in \leq ~\land~ (y,z) \in \leq  ~\Rightarrow~ (x,z) \in \leq$ and antisymmetry $(x,y) \in \leq \iff (y,x) \notin \leq$ hold $\forall x, y, z \in T$.
\end{definition}

Patterns as posets allow us to reuse and simplify the outputs from the discovery algorithm. Consider the inclusion of a DCR pattern describing a  sequential composition from $a$ to $b$ (similar to the flow construct in BPMN). A DCR model that captures a sequential behaviour will require 4 constraints in DCR: $\{a \conditionrel b, a\responserel b, a \excluderel a, b \excluderel b\}$. Consider $T=\{T_1:\textit{Condition}, T_2:\textit{Response}, T_3:\textit{Exclusion}, T_4:\textit{Sequence}\}$. The pattern poset $\langle T, \{(T_4, T_1), (T_4,T_2), (T_4,T_3)\}\rangle$ defines the dependency relations for a miner capable of mining sequential patterns. Additional patterns (e.g. exclusive choices, escalation patterns, etc), can be modelled in the same way. Pattern posets are finite, thus there exists minimal elements. 
%
\begin{algorithm2e}[t]
	\caption{Mining of rules starting from the extended DFG \label{alg:miner:pattern}}
	\footnotesize
	\DontPrintSemicolon
	\KwIn{$\langle T, \leq \rangle$: Pattern poset,
	    $G_X = (V, R, X)$: extended DFG}
	\DontPrintSemicolon
	\BlankLine
	$P \gets \langle V, M_\textit{init}, 
	\conditionrel=\emptyset, \responserel=\emptyset,
	 \includerel=\emptyset, \excluderel=\emptyset \rangle$ \Comment{Initial DCR graph}
	 $Rels \gets \emptyset$\\
    $CompRels \gets \emptyset$\\
    \ForEach(\Comment*[h]{Baseline for atomic patterns}){$t \in \textit{MinimalElements}(\langle T, \leq \rangle)$}{
        $\textit{Rels} \gets \textit{Rels} \cup  \textit{MineAtomic}(G_X, t)$\\
    }
    \ForEach(\Comment*[h]{Composite case}){$t \in T\backslash \textit{MinimalElements}(\langle T \leq \rangle)$}{
    	    $CompRels \gets CompRels \cup \textit{MineComposite}(G_X, t, Rels)$
    }
    \uIf{$CompRels \neq \emptyset$}{
    $P \gets P \oplus \textit{CompRels}$
    }
    \Else{$P \gets P \oplus \textit{Rels}$}
    \Return{$\textit{RemoveRedundancies}(P)$} \Comment{Apply  transitive reduction}
\end{algorithm2e}
\begin{algorithm2e}[t]
	\caption{Atomic miner \label{alg:miner:patternAtomic}}
	\footnotesize
	\DontPrintSemicolon
	\KwIn{$G_X = (V, R, X)$: extended DFG, $u$: DCR Pattern }
	\DontPrintSemicolon
	\BlankLine
	$\textit{Rels} \gets \emptyset$ \Comment{Empty dictionary of mined relations}
\ForEach{$(s,t) \in R$}{
\Comment{Pattern match with each atomic pattern}
    \Switch{u}{
        \Case{\textit{RESPONSE}}{
    	    \If{$X(s,\textsf{avgIdx}) < X(t,\textsf{avgIdx})$} {
    	        $\textit{Rels}[u] \gets \textit{Rels}[u] \cup (s,t,\responserel) $
    	    }        
        }
        \uCase{\textit{CONDITION}}{
    	    \If{$X(s, \textsf{avgFO}) < X(t, \textsf{avgFO}) \wedge X(s,\textsf{noTraceApp}) \geq X(t,\textsf{noTraceApp})$}{
    	        $\textit{Rels}[u] \gets \textit{Rels}[u] \cup (s,t,\conditionrel) $
    	        }
        }
        \Case{\textit{SELFEXCLUDE}}{
                \If{$X(s,noOccur) = 1$}{
    	        $\textit{Rels}[u] \gets \textit{Rels}[u] \cup (s,s,\excluderel) $
    	        }
        }
    }
	    \Comment{Further patterns here...}
	    \Return{\textit{Rels}}
	}
\end{algorithm2e}
\begin{algorithm2e}[t]
	\caption{Composite miner \label{alg:miner:patternComposite}}
	\footnotesize
	\DontPrintSemicolon
	\KwIn{$G_X = (V, R, X)$: extended DFG, $u$: DCR Pattern, $Rels \colon$ Mined Relations}
	\DontPrintSemicolon
	\BlankLine
    \Switch{u}{
        \uCase{\textit{EXCLUDEINCLUDE}}{
            \Return{$\textit{Rels}[\textit{SELFEXCLUDE}]\ \cup \textit{Rels}[\textit{PRECEDENCE}] \cup \textit{Rels}[\textit{NOTCHAINSUCCESION}]$} \Comment{Removes redundant relations}
        }
	    \Comment{Further patterns here}
	}
\end{algorithm2e}
The generation of a DCR model from an extended DFG is described in Algorithm~\ref{alg:miner:pattern}. We illustrate the mining of DCR \textit{conditions}, \textit{responses} and \textit{self-responses}, but more patterns are available in~\cite{starklit2021a}. The algorithm takes as input an extended DFG $G_X$ and a pattern poset. It starts by creating an empty DCR graph $P$ with activities equal to the nodes in $G_X$ and initial marking $M_{\textit{init}} = \{\emptyset,\emptyset, V\}$, that is, all events are included, not pending and not executed. 
We then split the processing between atomic patterns (those with no dependencies) and composite patterns. The map $\textit{Rel}$ stores the relations from atomic patterns, that will be used for the composite miner. We use the merge notation $P \oplus \textit{Rels}$ to denote the result of the creation of a DCR graph whose activities and markings are the same as $P$, and whose relations are the pairwise union of the range of $\textit{Rels}$ and its corresponding relational structure in $P$.
Line 8 applies a transitive reduction strategy~\cite{Back2021}, reducing the number of relations while maintaining identical reachability properties.

The atomic and composite miners are described in Algorithms~\ref{alg:miner:patternAtomic}, and \ref{alg:miner:patternComposite}. The atomic miner in  Algorithm~\ref{alg:miner:patternAtomic} iterates over all node dependencies in the DFG and the pattern matches with the existing set of implemented patterns. Take the case of response. We will identify a response if the average occurrence of $s$ is before $t$ (line 6). This condition, together with the dependency between $s$ and $t$ in $G_X$ is sufficient to infer a response constraint from $s$ to $t$.
To detect conditions, the algorithm verifies a different set of properties: given a DFG dependency  between $s$ and $t$, it checks that the first occurrence of $s$ precedes $t$ and that $s$ and $t$ appeared in the same traces  (approximated by counting the number of traces containing both activities, line 9). The atomic pattern refers to self exclusions identified via the  occurrences of an activity in the stream. Other atomic patterns include precedence, or absence of chain successions referred to in Algorithm \ref{alg:miner:patternComposite}.

The composite miner receives the DFG, a pattern, and the list of mined relations from atomic patterns. We provide an example for the case of include and exclude relations. This pattern is assembled as the combination of self-exclusions, precedence, and not chain successions. As these atomic patterns generate each a set of include/exclude relations, the pattern just takes the set union construction.

\paragraph{Suitability of the Algorithms for Streaming Settings.}
Whenever discussing algorithms that can tackle the streaming process mining problem~\cite{DBLP:reference/bdt/Burattin19}, it is important to keep in mind that while a stream is assumed to be infinite, only a finite amount of memory can be used to store all information and that the time complexity for processing each event must be constant. Concerning the memory, an upper bound on the number of stored events in Alg.~\ref{alg:meta} is given by $m_t \cdot m_e$ which happens when each trace contains more than $m_e$ events and at least $m_t$ traces are seen in parallel. Additionally, please note that the extended DFG is also finite since there is a node for each activity contained in the memory.
Concerning the time complexity, Alg.~\ref{alg:meta} does not perform any unbounded backtracking but, instead, for each event, it operates using just maps that have amortized constant complexity or on the extended DFG (which has finite, controlled size). The same observation holds for Alg.~\ref{alg:miner:pattern} as it iterates on the extended DFG which has a size bounded by the provided parameters (and hence, can be considered constant).



\section{Experimental Evaluation} \label{sec:validity}

To validate the result of the approach presented in the paper we executed several tests, first to validate quantitatively the streaming discovery on synthetic data, then to qualitatively  evaluate the whole approach on a real dataset.

\subsection{Quantitative Evaluation of Streaming Discovery}

\begin{figure}
    \centering
    \begin{subfigure}{\textwidth}
        \includegraphics[width=.5\textwidth]{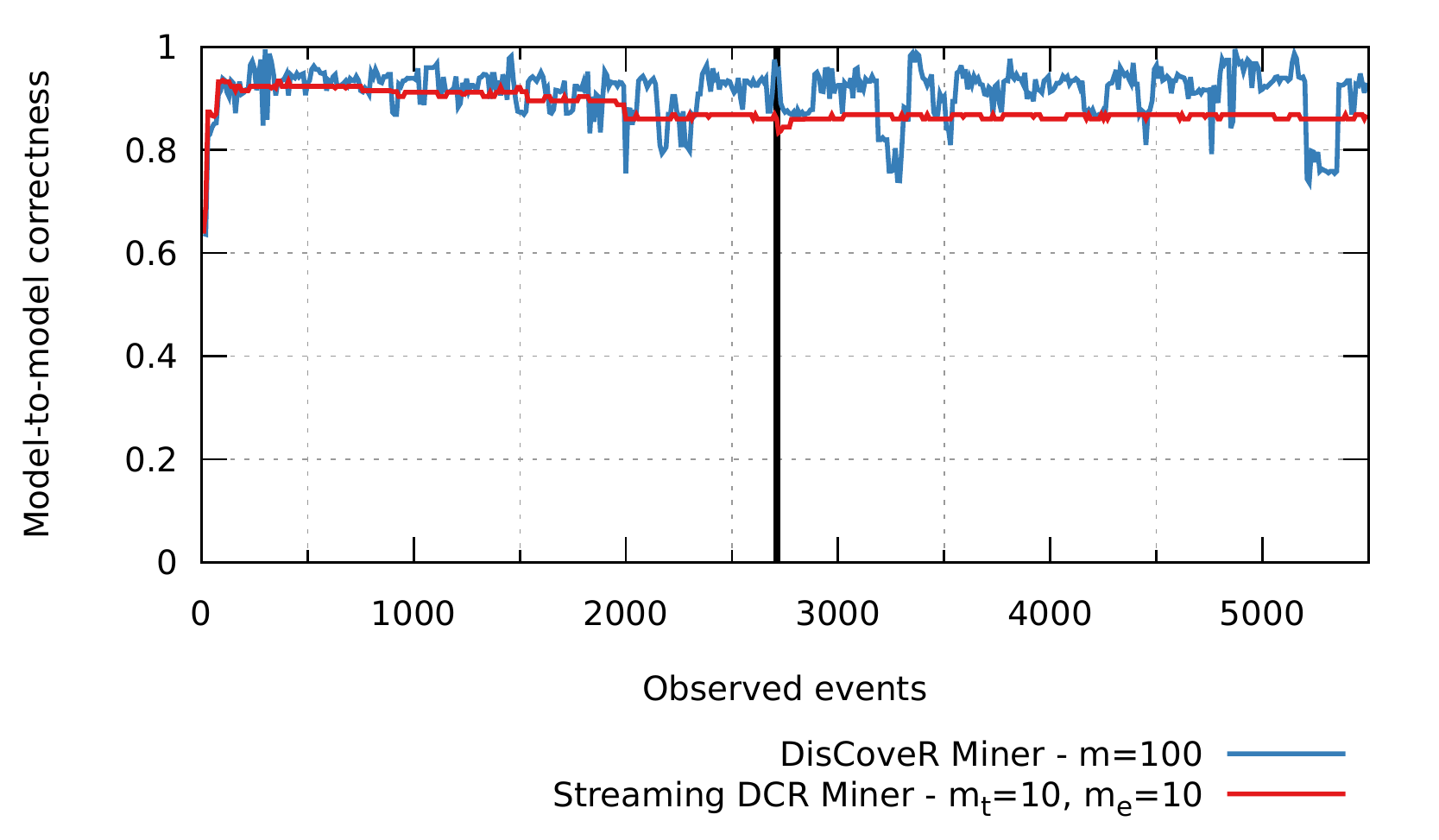}%
        \includegraphics[width=.5\textwidth]{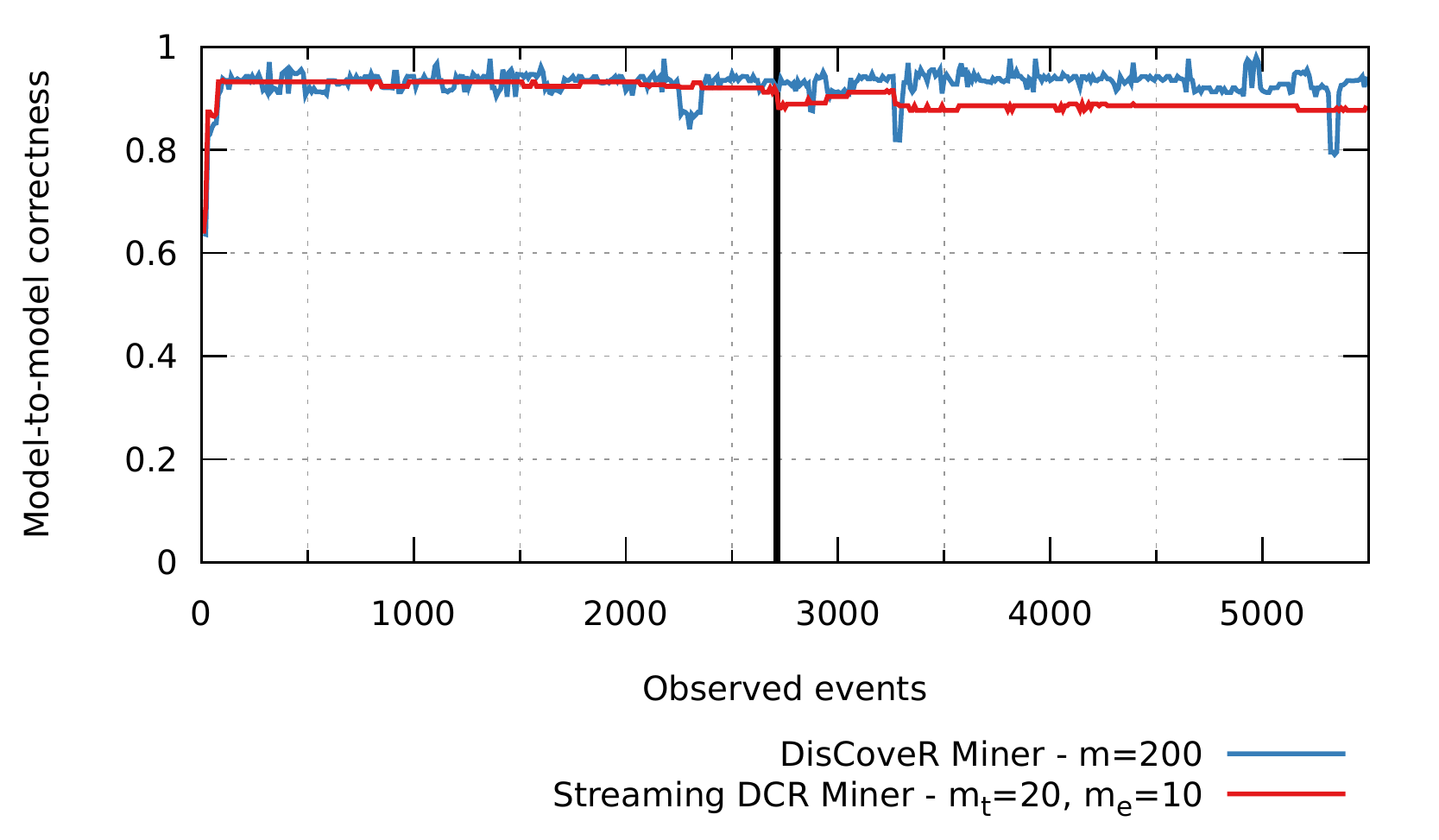}
        \includegraphics[width=.5\textwidth]{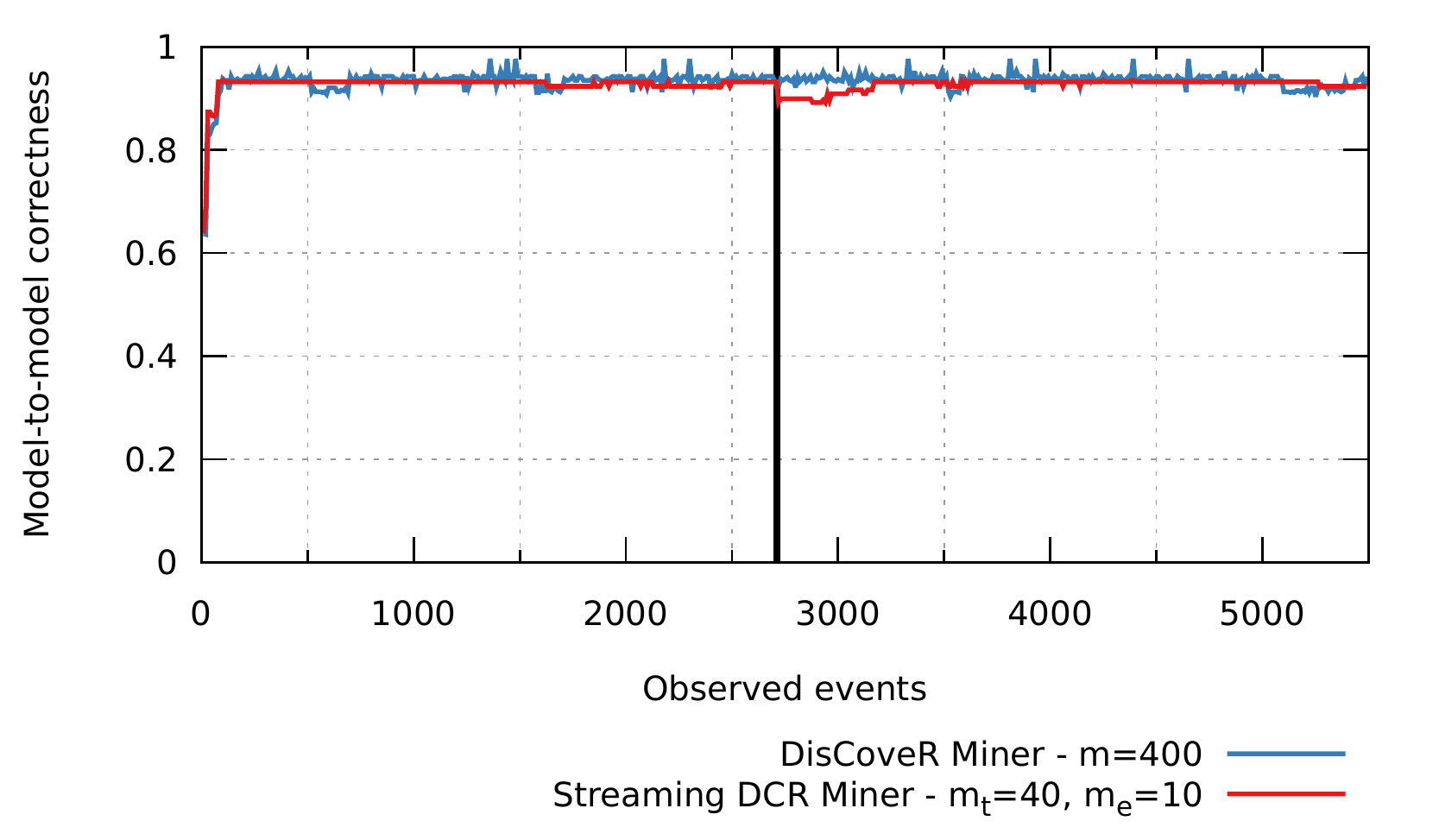}%
        \includegraphics[width=.5\textwidth]{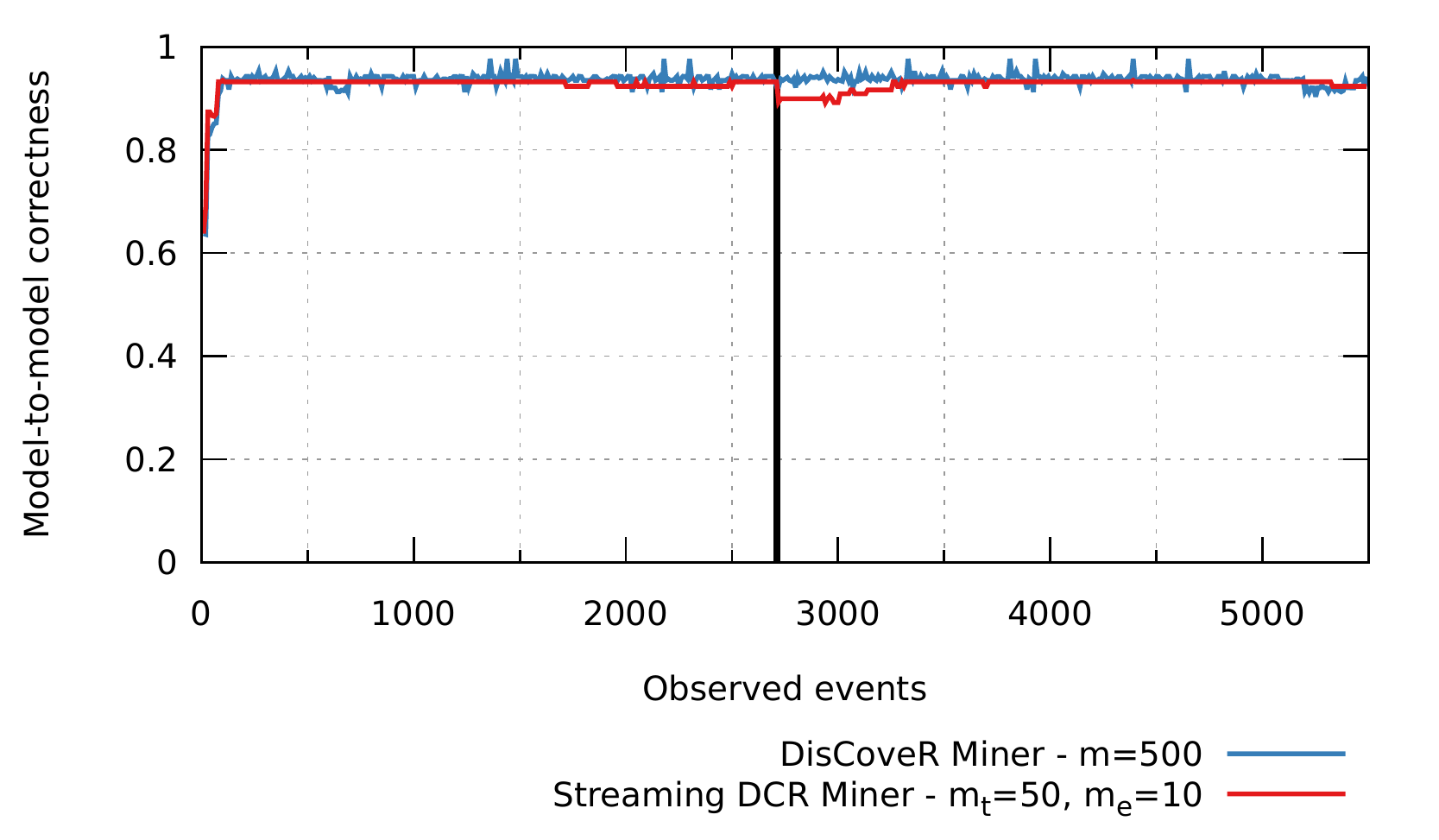}
        
        \caption{Performance comparison on a simple stream.}
        \label{fig:simple-stream} 
    \end{subfigure}
    \begin{subfigure}{\textwidth}
        \includegraphics[width=.5\textwidth]{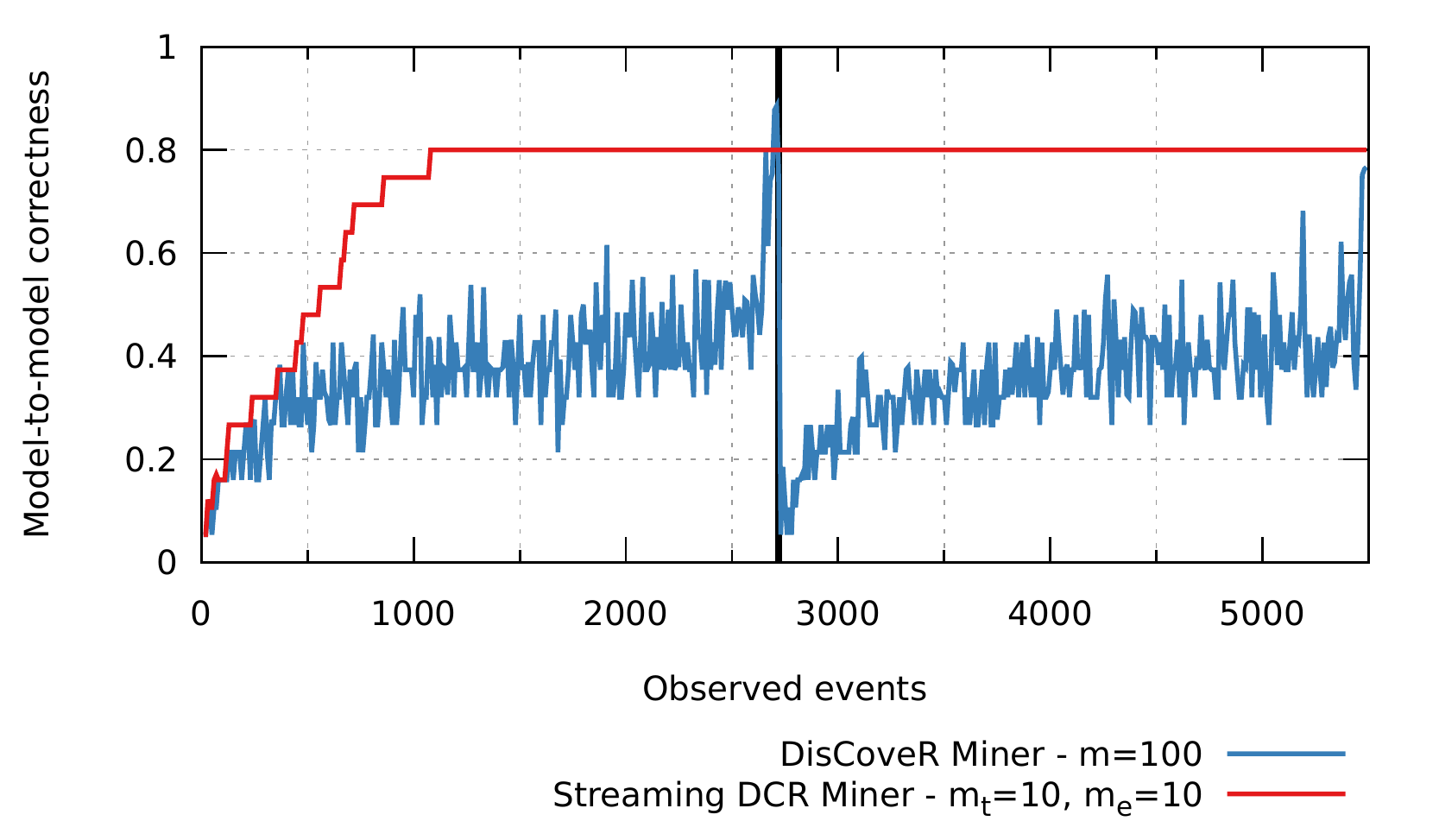}%
        \includegraphics[width=.5\textwidth]{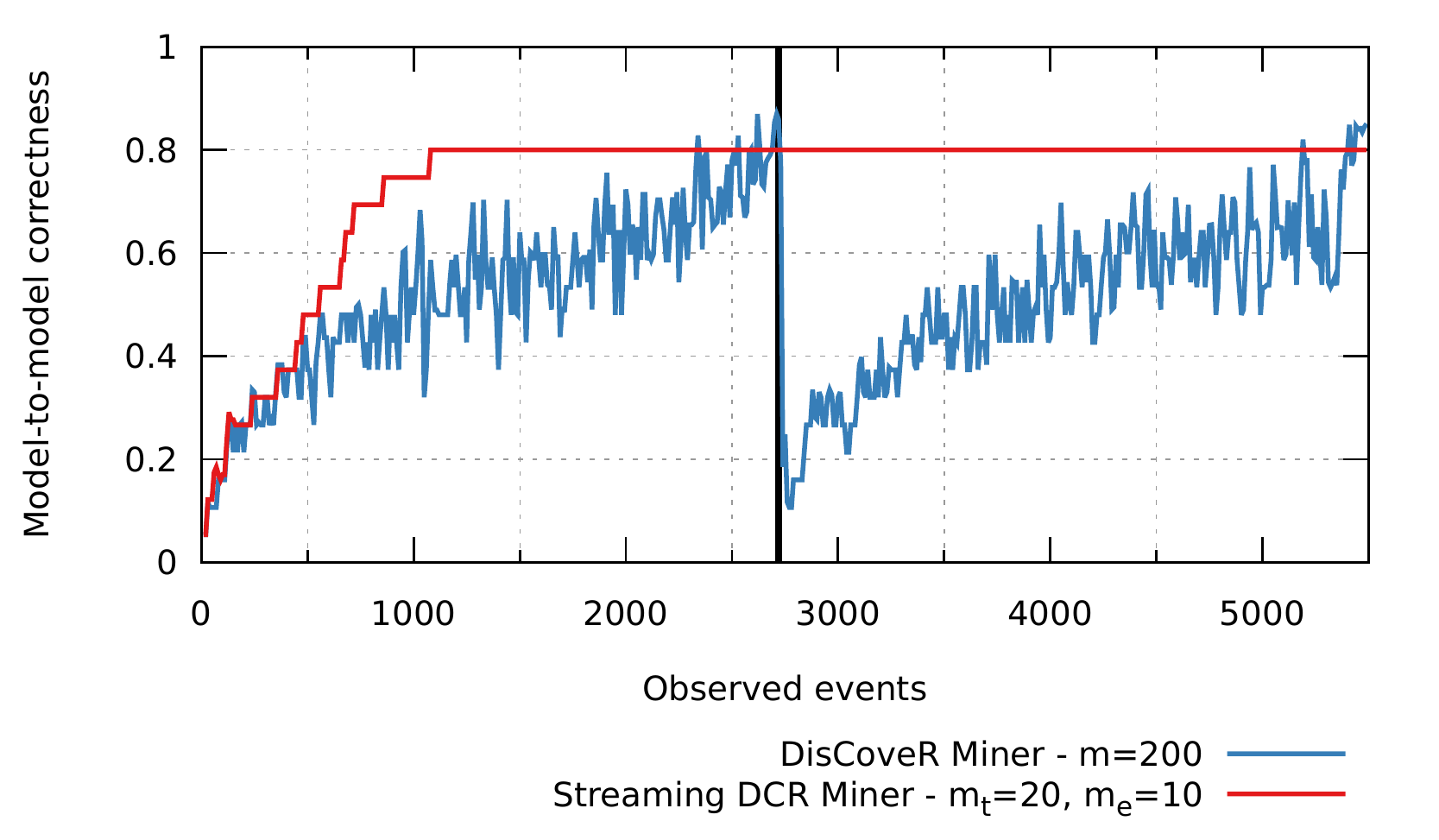}
        \includegraphics[width=.5\textwidth]{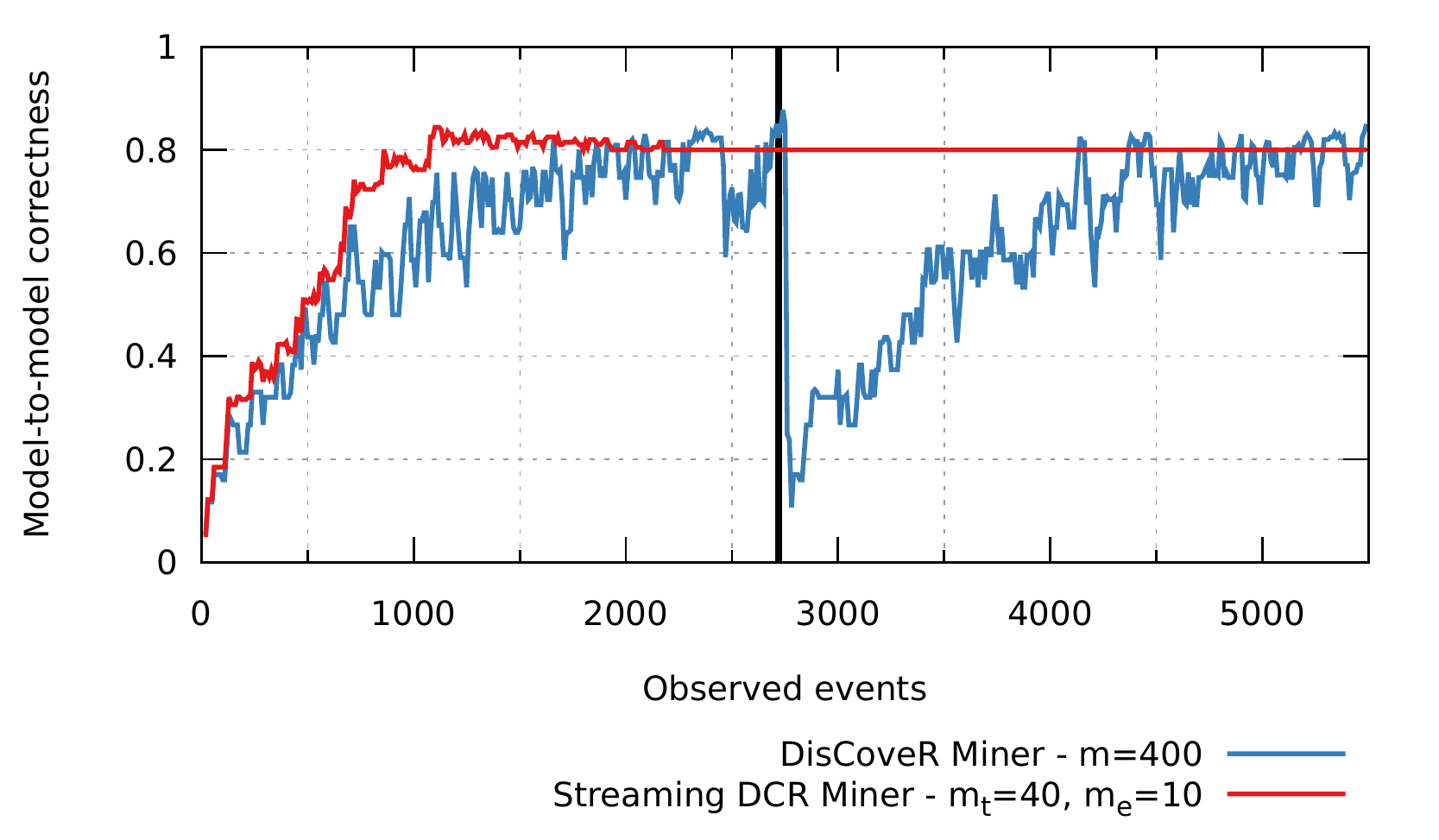}%
        \includegraphics[width=.5\textwidth]{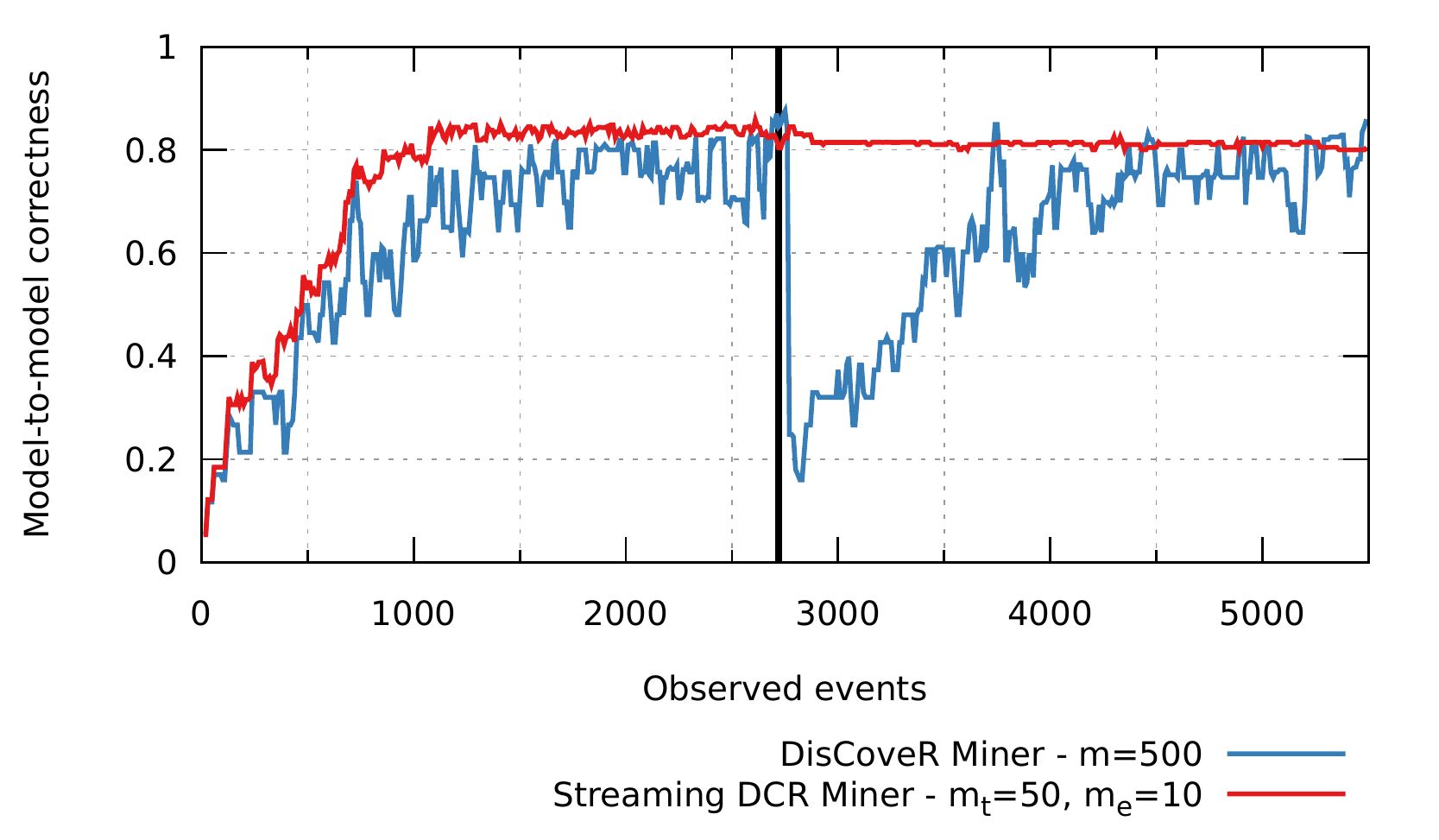}
        \caption{Performance comparison on a complex stream.}
        \label{fig:complex-stream} 
    \end{subfigure}
    \caption{Performance comparison between the offline DisCoveR miner and the streaming DCR Miner with the same amount of storage available  (with a capacity of up to 100, 200, 400 and 500 events). A vertical black bar indicates a drift in the model generating the stream.}
    \label{fig:synt}
\end{figure}

Our goal with the first quantitative evaluation is to compare the stability of the streaming DCR miner against sudden changes. We compare with other process discovery algorithms for DCR graphs, in this case, the DisCoveR miner~\cite{Back2021}. The tests are performed against a publicly available dataset of events streams~\cite{ceravolo2020evaluation}. This dataset includes (1)~a synthetic stream inspired by a loan application process, and (2)~perturbations to the original stream using change patterns~\cite{weber2008change}. Recall that the DisCoveR miner is an \emph{offline} miner, thus it assumes an infinite memory model. To provide a fairer evaluation we need to parameterize DisCoveR with the same amount of available memory. 
We divided the experiment into two parts: a simple stream where the observations of each process instance arrive in an ordered manner (i.e., one complete process instance at a time) and a complex stream where observations from many instances arrive intertwined. As no initial DCR graph exists for this process, we used the DisCoveR miner in its original (offline) setting to generate a baseline graph using the entire dataset.
This model (the one calculated with offline DisCoveR) was used to calculate the model-to-model similarity between the DCR stream miner and the DisCoveR miner with memory limits. For the sake of simplicity, in this paper, we considered only the case of sudden drifts, while we discuss other types of drift in future work.

In order to quantify the extent of the similarity between the baseline model and the discovered one, we developed a model-to-model metric capable of quantifying to which extent two DCR graphs are similar. This metric can be used, for example, to identify which process is currently being executed with respect to a repository of candidate processes, or by quantifying the \emph{change rate} of the same process over time.
The metric takes as input two DCR graphs $P$ and $Q$ as well as a weight relation $W$ that associates each DCR relation in $\phi$ (cf. Def.~\ref{def:dcr}) with a weight, plus one additional weight for the activities. Then it computes the weighted Jaccard similarity~\cite{jaccard} of the sets of relations and the set of activities, similarly to what happens in \cite{AiolliBS11} imperative models:
\begin{definition}[DCR Model-to-Model metric]
    Given $P$ and $Q$ two DCR graphs, and $W: \phi \cup \{\textsf{act}\} \to \mathbb{R}$ a weight function  in the range $[0,1]$ such that $\sum_{r \in \phi \cup \{\textsf{act}\}} W(r)=1$. The model-to-model similarity metric is defined as:
    \begin{equation}
        S(P,Q,W) = W(\textsf{act}) \cdot \frac{|P_\mathcal{A} \cap Q_\mathcal{A}|}{|P_\mathcal{A} \cup Q_\mathcal{A}|} +  \sum_{r \in \phi} W(r) \cdot \frac{|P_r \cap Q_r |}{|P_r \cup Q_r|}
    \end{equation}
\end{definition}

The model-to-model metric is a similarity metric comparing the relations in each of the two DCR graphs, thus returning a value between 0 and 1, where 1 indicates a perfect match and 0 stands for no match at all. A brief evaluation of the metric is reported in Appendix~\ref{sec:appendix:eval}.

The results of the quantitative evaluation are reported in Fig.~\ref{fig:simple-stream} and~\ref{fig:complex-stream} (for simple and complex stream resp.). Each figure shows the performance of the incremental version of DisCoveR and the streaming DCR miner against four different configurations over time. The vertical black bars indicate where a sudden drift occurred in the stream. While the performance for the simple stream is very good for both the DisCoveR and the streaming DCR miners, when the stream becomes more complicated (i.e., Fig.~\ref{fig:complex-stream}), DisCoveR becomes less effective, and, though its average performance increases over time, the presence of the drift completely disrupt the accuracy. In contrast, our approach is much more robust to the drift and much more stable over time, proving its ability at managing the available memory in a much more effective way.

\subsection{Qualitative Evaluation of the Entire Approach}

For the final evaluation, we run a qualitative analysis on a real stream. We compared the results of the streaming miner against a real process model from one of our partner companies: the Dreyers Fond case~\cite{debois2015analysis}. The stream contains activities for a grant application system from December 2013 until May 2015, with 6,470 events, some of which belong to partial (never finished) traces. The data was generated from a DCR graph\footnote{The latest version is available at \url{https://www.dcrgraphs.net/tool/main/Graph?id=59a932f8-1011-4232-bbc5-9b39efb1fc18}.}.  We used this data to explore whether the streaming miner is capable to identify actual drifts and validate such drifts with the process owner. We finally studied the  computational overhead on data from a real scenario and the number of constraints identified in a realistic case. 

\begin{figure}
    \centering
    \begin{subfigure}[b]{\textwidth}
        \includegraphics[width=\textwidth]{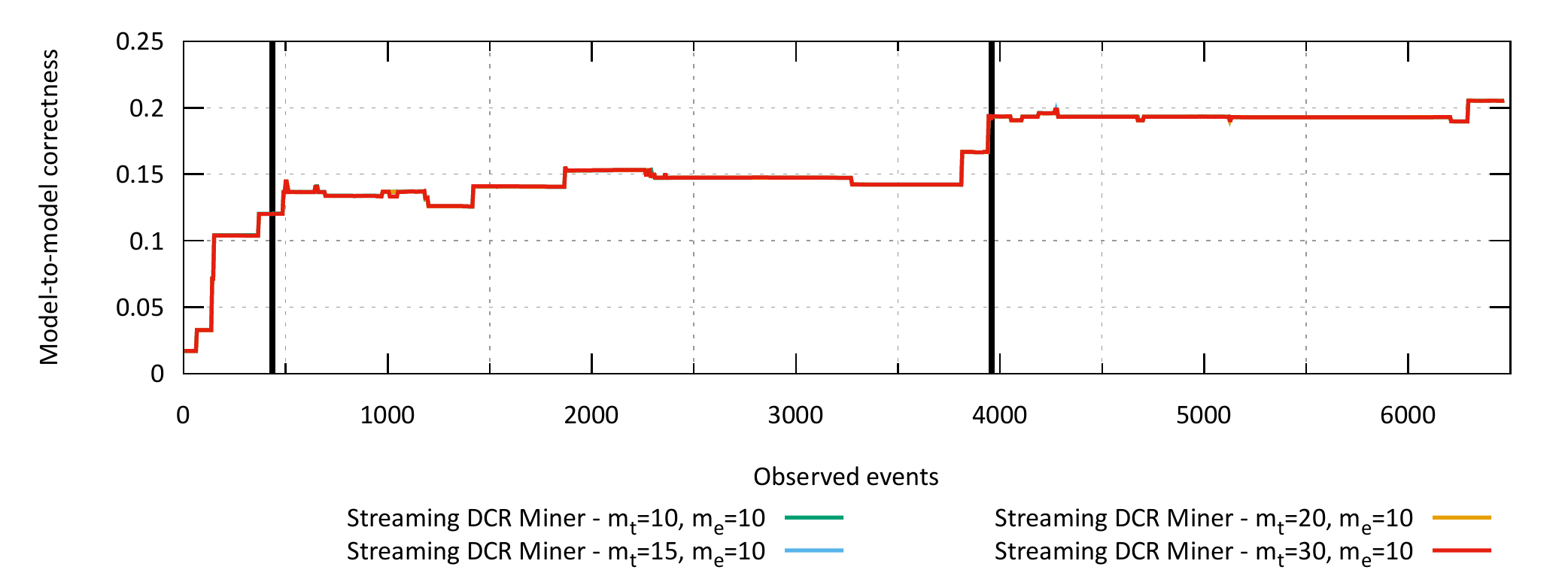}
        \caption{Model-to-model similarity calculated between the model mined at each point in time and the normative model, with different memory configurations. The vertical black bars are positioned according to actual drifts in the process. All lines are overlapped.}
        \label{fig:dreyers:jaccard}
    \end{subfigure}
    \begin{subfigure}[b]{\textwidth}
        \includegraphics[width=\textwidth]{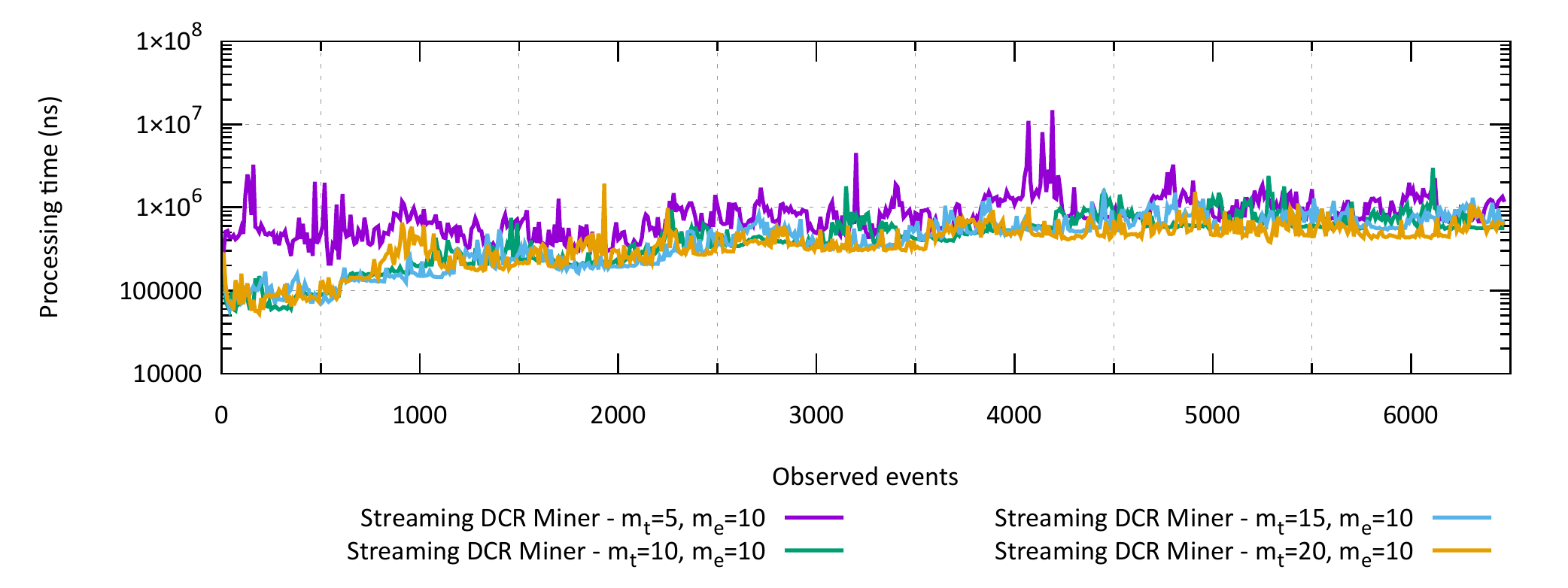}
        \caption{Time required to process each event (expressed in nanoseconds, log scale) over time on the real stream data, with four different memory configurations.}
        \label{fig:dreyers:time}
    \end{subfigure}
    \begin{subfigure}[b]{\textwidth}
        \includegraphics[width=\textwidth]{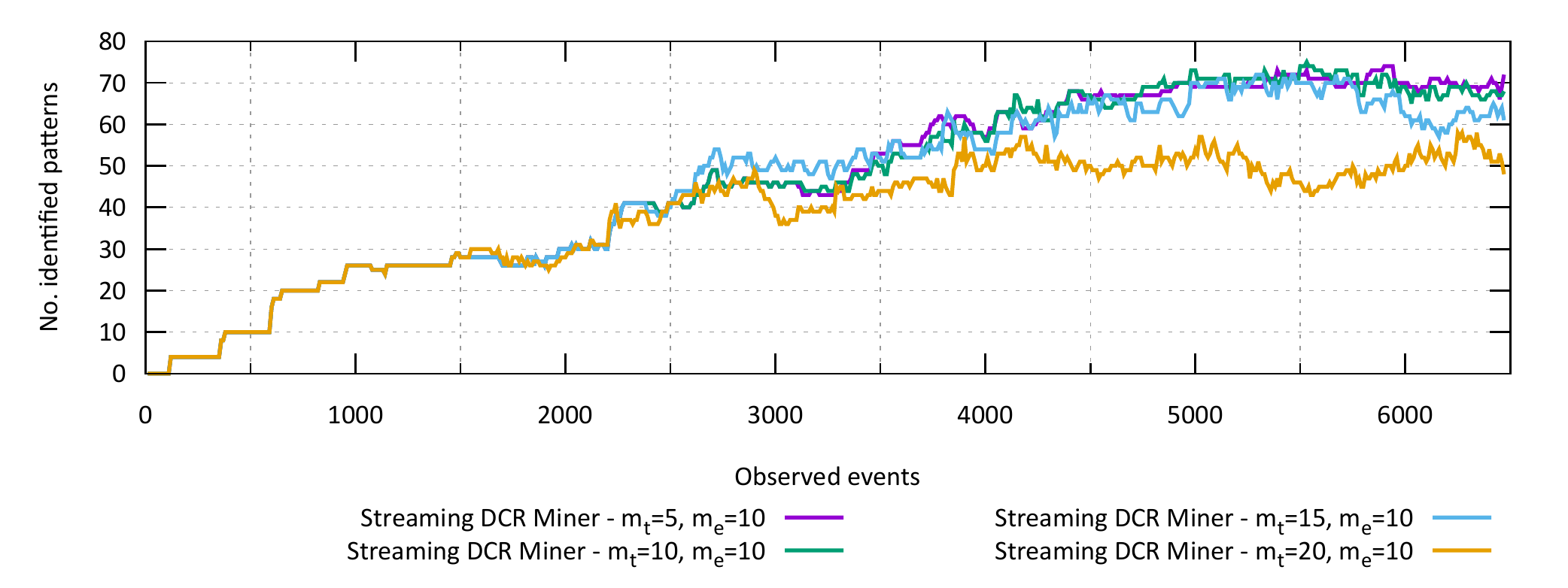}
        \caption{Number of constraints extracted over time on the real stream data, with four different memory configurations.}
        \label{fig:dreyers:constraints}
    \end{subfigure}
    \caption{Analyses on the real stream referring to the Dreyers Fond.}
    \label{fig:dreyersfond}
\end{figure}

Fig.~\ref{fig:dreyers:jaccard} presents the model-to-model similarity against the normative model. It is important to point out that the normative model includes language features that decrease the number of constraints in the model (e.g. nesting~\cite{DBLP:conf/fsen/HildebrandtMS11} and data events), resulting in a low absolute similarity score. However, two important sudden drifts in the model were detected: one before March 2013, and another in May 2015 (highlighted in the figure as well). We inquired the process owner regarding these changes, and they were both confirmed. In the first case, a testing phase, before the process entered into production (cit. \emph{``The Dreyers Fond went live in December 2013 but in reality, did not process any applications until March 2014''}), took place. The second drift uncovered a system malfunction (cit. \emph{``In May 2015 I recall we had a server crash where we manually had to fix things''}). In addition, we enquired regarding the smaller changes during the lifetime of the process (resulting in faint model-to-model similarity changes in \ref{fig:dreyersfond}), according to the process owner, such changes represent seasonal drifts: the process starts receiving applications first, and it gradually changes into observing events regarding their application reviews.  

In  the second analysis, depicted in Fig.~\ref{fig:dreyers:time}, we reported the time required to process each event. As the graph shows, processing each event on a consumer laptop (Macbook) requires about 1 millisecond, thus showing the applicability of our technique in real settings.

Finally, Fig.~\ref{fig:dreyers:constraints} shows the number of discovered constraints over time. It is worth noticing that initially, the number of constraints grows consistently for all different configurations, however, after a while, the configuration with the larger memory extracts a smaller number of constraints. This is because more memory available means having more observations and thus more potential counterexamples to the requirements for having a constraint (cf. Alg.~\ref{alg:miner:pattern}).

\subsection{Discussion}

One of the limitations of the approach regards precision with respect to offline miners. A limiting aspect of the approach relies on the choice of the intermediate structure. As recently pointed out in~\cite{dfg_aalst,DBLP:journals/corr/abs-2202-08314}, a DFG representation may report confusing model behavior as it simplifies the observations using purely a frequency-based threshold. A DFG is in essence an imperative data structure that captures the most common flows that appear in a stream. This, in a sense, goes against the declarative paradigm as a second-class citizen with respect to declarative constraints. We hypothesize that the choice of the DFG as an intermediate data structure carries out a loss of precision with respect to the DisCoveR miner in offline settings. However, in an online setting, the DFG still provides a valid approximation to observations of streams where we do not have complete traces. This is far from an abnormal situation: IoT communication protocols such as MQTT~\cite{hunkeler2008mqtt} assume that subscriber nodes might connect to the network \emph{after} the communications have started, not being able to identify starting nodes.
Specifically, in a streaming setting it is impossible to know exactly when a certain execution is complete and, especially in declarative settings, certain constraints describe liveness behaviors that can only be verified  after a whole trace has been completely inspected. While watermarking techniques~\cite{osti_1823361} could be employed to cope with \textit{lateness} issues, we have decided to favor  self-contained approaches in this paper, leaving for future work the exploration of watermarking techniques.

\section{Conclusion and Future Work} \label{sec:conclusions}

This paper presented a novel streaming discovery technique capable of extracting declarative models, expressed using the DCR language, from event streams. Additionally, a model-to-model metric is reported which allows understanding if and to what extent two DCR models are the same. A thorough experimental evaluation, comprising both synthetic and real data, validated the two contributions separately as well as their combination in a qualitative fashion, which included interviews with the process owner.

We plan to explore several directions in future work. Regarding the miner, we plan to extend its capabilities to the identification of sub-processes, nesting, and data constraints. Regarding the model-to-model similarity, we would like to embed more semantic aspects, such as mentioned in~\cite{DBLP:conf/fase/LopezDSH20}. A possible limitation of the streaming miner algorithm approach followed here relates to the updating mechanism. Currently lines 22--24 of Algorithm~\ref{alg:meta} perform updates based entirely on periodic updates triggered by time, which will generate notifications even when no potential changes in the model have been identified. A possibility to extend the algorithm will be to integrate the model-to-model similarity as a parameter to the discovery algorithm, so models only get updated after a given change threshold (a similarity value specified by the user) is reached. 

\paragraph{Acknowledgments} We would like to thank Morten Marquard from DCR Solutions for providing valuable information regarding the Dreyers Fond case.


\bibliographystyle{splncs04}
\bibliography{references,references-andrea}

\appendix

\section{Quantitative Evaluation of Model-to-model Metric}
\label{sec:appendix:eval}

\begin{figure}[t]
    \centering
    \includegraphics[width=\textwidth]{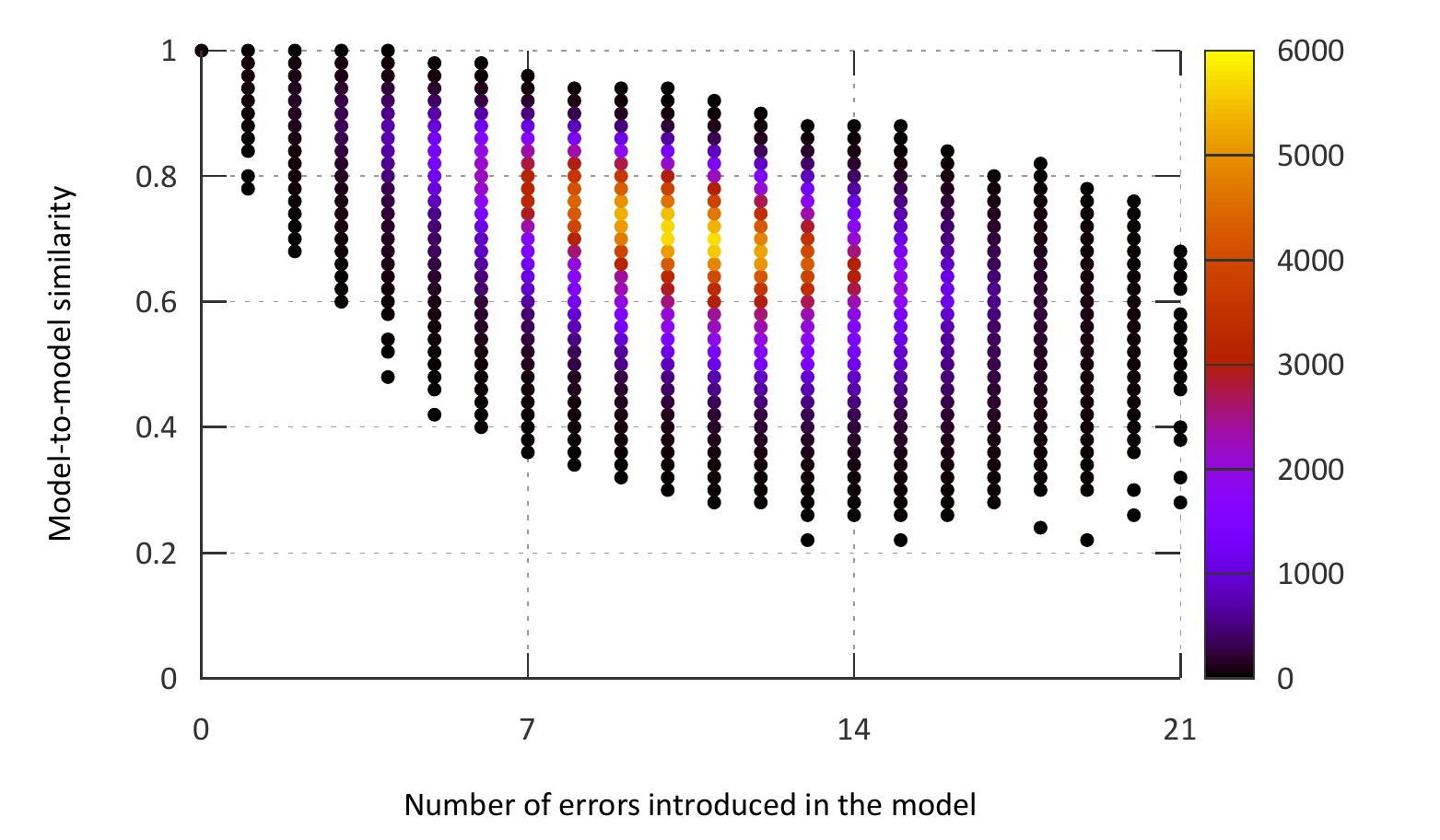}
    \caption{Scatter plot showing the correlation between the model-to-model metric and the number of changes introduced in the model. The colour indicates the density of observations.}
    \label{fig:m2m}
\end{figure}

To validate the quality of the model-to-model metric, we used a dataset of 28 DCR process models collected from previous mapping efforts~\cite{DBLP:conf/caise/LopezSNM21} and, for each model, we randomly introduced variations such as: adding new activities connected to the existing fragments, adding disconnected activities, deleting existing activities (with corresponding constraints), adding constraints, removing constraints, and swapping activity labels in the process. By systematically applying all possible combinations of variations in a different amount (e.g., adding 1/2/3 activities and nothing else; adding 1/2/3 activities and removing 1/2/3 constraints) we ended up with a total of 455,826 process models with a quantifiable amount of variation from the 28 starting processes.

Fig.~\ref{fig:m2m} shows each variation on a scatter plot where the $x$ axis refers to the number of variations introduced in the model and the $y$ axis refers to the model-to-model similarity. The color indicates the number of models in the proximity of each point (since multiple processes have very close similarity scores). For identifying the optimal weights we solve an optimization problem, aiming at finding the highest correlation between the points, ending up with: $W = \{
(\conditionrel, 0.06),
(\responserel, 0.07),
(\milestonerel, 0.06),
(\includerel, 0.07),
(\excluderel, 0.13),
(\textsf{act}, 0.61)
\}$ which leads to a Pearson's correlation of -0.56 and a Spearman's correlation of -0.55. These values indicate that our metric is indeed capable of capturing the changes. As the metric is very compact (value in $[0,1]$) and operates just on the topological structure of the model, it cannot identify all details. However, the metric benefits from the fast speed of computation.

\end{document}